\documentclass{article}

\usepackage{graphicx}

\title{Node discovery problem for a social network}
\author{Yoshiharu Maeno\footnote{Yoshiharu Maeno, Ph.D. is a founder management consultant of Social Design Group, Sengoku 1-6-38F, Bunkyo-ku, Tokyo 112-0011 Japan. Telephone: +81-90-8009-1968. Email: maeno.yoshiharu@socialdesigngroup.com.} \\ Social Design Group}

\date{}

\begin{document}

\maketitle

\begin{abstract}
Methods to solve a node discovery problem for a social network are presented. Covert nodes refer to the nodes which are not observable directly. They transmit the influence and affect the resulting collaborative activities among the persons in a social network, but do not appear in the surveillance logs which record the participants of the collaborative activities. Discovering the covert nodes is identifying the suspicious logs where the covert nodes would appear if the covert nodes became overt. The performance of the methods is demonstrated with a test dataset generated from computationally synthesized networks and a real organization.
\end{abstract}

\newpage

\twocolumn

\section{Introduction}
\label{Introduction}

Covert nodes refer to persons who transmit the influence and affect the resulting collaborative activities among the persons in a social network, but do not appear in the surveillance logs which record the participants of the activities. The covert nodes are not observable directly. It aids us in discovering and approaching to the covert nodes to identify the suspicious surveillance logs where the covert nodes would appear if they became overt. I call this problem a node discovery problem for a social network.

Where do we encounter such a problem? Globally networked clandestine organizations such as terrorists, criminals, or drug smugglers are great threat to the civilized societies \cite{Sag04}. Terrorism attacks cause great economic, social and environmental damage. Active non-routine responses to the attacks are necessary as well as the damage recovery management. The short-term target of the responses is the arrest of the perpetrators. The long-term target of the responses is identifying and dismantling the covert organizational foundation which raises, encourages, and helps the perpetrators. The threat will be mitigated and eliminated by discovering covert leaders and critical conspirators of the clandestine organizations. The difficulty of such discovery lies in the limited capability of surveillance. Information on the leaders and critical conspirators are missing because it is usually hidden by the organization intentionally.

Let me show an example in the 9/11 terrorist attack in 2001 \cite{Kre02}. Mustafa A. Al-Hisawi, whose alternate name was Mustafa Al-Hawsawi, was alleged to be a wire-puller who had acted as a financial manager of Al Qaeda. He had attempted to help terrorists enter the United States, and provided the hijackers of the 4 aircrafts with financial support worth more than 300,000 dollars. Furthermore, Osama bin Laden is suspected to be a wire-puller behind Mustafa A. Al-Hisawi and the conspirators behind the hijackers. These persons were not recognized as wire-pullers at the time of the attack. They were the nodes to discover from the information on the collaborative activities of the perpetrators and conspirators known at that moment.

In this paper, I present two methods to solve the node discovery problem. One is a heuristic method in \cite{Mae09}, which demonstrates a simulation experiment of the node discovery problem for the social network of the 9/11 perpetrators. The other is a statistical inference method which I propose in this paper. The method employs the maximal likelihood estimation and an anomaly detection technique. Section \ref{Problem} defines the node discovery problem mathematically. Section \ref{Solution} presents the two methods. Section \ref{Test} introduces the test dataset generated from computationally synthesized networks and a real clandestine organization. Section \ref{Performance} demonstrates the performance characteristics of the methods (precision, recall, and van Rijsbergen's F measure \cite{Kor97}). Section \ref{Conclusion} presents the issues and future perspectives as concluding remarks. Section \ref{Related Work} summarizes the related works.

\section{Related Work}
\label{Related Work}

The social network analysis is a study of social structures made of nodes which are linked by one or more specific types of relationship. Examples of the relationship are influence transmission in communication or presence of trust in collaboration \cite{Lav07}. Network topological characteristics of clandestine terrorist organizations \cite{Kre02} and criminal organizations \cite{Kle02} are studied. Trade-off between staying secret and efficiently securing coordination and control is of particular interest \cite{Mor07}. The impact of the network topology to the trade-off is analyzed \cite{Lin09}.

Research interests have been moving from describing organizational structure to discovering dynamical phenomena on a social network. A link discovery predicts the existence of an unknown link between two nodes from the information on the known attributes of the nodes and the known links \cite{Cla08}. It is one of the tasks of link mining \cite{Get05}. The link discovery techniques are combined with domain-specific heuristics. The collaboration between scientists can be predicted from the published co-authorship \cite{Lib04}. The friendship between people is inferred from the information available on their web pages \cite{Ada03}.

Markov random network is a model of the joint probability distribution of random variables. It is an undirected graphical model similar to a Bayesian network. The Markov random network is used to learn the dependency between the links which shares a node. The Markov random network is one of the dependence graphs \cite{Fra86}, which models the dependency between links. Extension to hierarchical models \cite{Laz99}, multiple networks (treating different types of relationships) \cite{Pat99}, valued networks (with nodal attributes) \cite{Rob99}, higher order dependency between the links which share no nodes \cite{Pat02}, and 2-block chain graphs (associating one set of explanatory variables with the other set of outcome variables) \cite{Rob01} are studied. A family of such extensions and model elaborations is named the exponential random graph \cite{And99}.

In addition to the link discovery, the related research topics are the exploration of an unknown network structure \cite{New07}, the discovery of a community structure \cite{Pal05}, the inference of a network topology \cite{Rab08}, the detection of an anomaly in a network \cite{Sil09}, and the discovery of unknown nodes \cite{Mae07}, \cite{Mae09}. Stochastic modeling to predict terrorism attacks \cite{Sin04} is relevant practically. The idea of machine learning of latent variables \cite{Sil06} is potentially applicable to discovering an unknown network structure. 

\section{Problem definition}
\label{Problem}

The node discovery problem is defined mathematically in this section. A node represents a person in a social network. A link represents a relationship which transmits the influence between persons. The symbols $n_{j} \ (j=0,1,\cdots)$ represent the nodes. Some nodes are overt (observable), but the others are covert (unobservable). $\mbox{\boldmath{$O$}}$ denotes the overt nodes; $\{n_{0}, n_{1}, \cdots, n_{N-1} \}$. Its cardinality is $|\mbox{\boldmath{$O$}}|=N$. $\mbox{\boldmath{$C$}} = \overline{\mbox{\boldmath{$O$}}}$ denotes the covert nodes; $\{n_{N},n_{N+1}, \cdots, n_{M-1} \}$. Its cardinality is $|\mbox{\boldmath{$C$}}|=M-N$. The whole nodes in a social network is $\mbox{\boldmath{$O$}} \cup \mbox{\boldmath{$C$}}$. The number of the nodes is $M$. The unobservability of the covert nodes arises either from a technical defect of surveillance means or an intentional cover-up operation. 

The symbol $\delta_{i}$ represent a set of participants in a particular collaborative activity. It is the $i$-th activity pattern among the nodes. A pattern $\delta_{i}$ is a set of nodes; $\delta_{i}$ is a subset of $\mbox{\boldmath{$O$}} \cup \mbox{\boldmath{$C$}}$. For example, the nodes in an collaborative activity pattern are those who joined a particular conference call. That is, a pattern is a co-occurrence among the nodes \cite{Rab08}. The unobservability of the covert nodes does not affect the activity patterns themselves.

A simple hub-and-spoke model is assumed as a model of the influence transmission over the links resulting the collaborative activities among the nodes. The way how the influence is transmitted governs the set of possible activity patterns $\{\delta_{i}\}$. The network topology and the influence transmission are described by some probability parameters. The probability where the influence transmits from an initiating node $n_{j}$ to a responder node $n_{k}$ is $r_{jk}$. The influence transmits to multiple responders independently in parallel. It is similar to the degree of collaboration probability in trust modeling \cite{Lav07}. The constraints are $0 \leq r_{jk}$ and $\sum_{k \neq j} r_{jk} \leq 1$. The quantity $f_{j}$ is the probability where the node $n_{j}$ becomes an initiator. The constraints are $0 \leq f_{j}$ and $\sum_{j=0}^{N-1} f_{j} = 1$. These parameters are defined for the whole nodes in a social network (both the nodes in $\mbox{\boldmath{$O$}}$ and $\mbox{\boldmath{$C$}}$).

A surveillance log $d_{i}$ records a set of the overt nodes in a collaborative activity pattern; $\delta_{i}$. It is given by eq.(\ref{Dataset2}). A log $d_{i}$ is a subset of $\mbox{\boldmath{$O$}}$. The number of data is $D$. A set $\{ d_{i} \}$ is the whole surveillance logs dataset.
\begin{eqnarray}
d_{i} = \delta_{i} \cap \mbox{\boldmath{$O$}} \ (0 \leq i < D).
\label{Dataset2}
\end{eqnarray}

Note that neither an individual node nor a single link alone can be observed directly, but nodes can be observed collectively as a collaborative activity pattern. The dataset $\{ d_{i} \}$ can be expressed by a 2-dimensional $D \times N$ matrix of binary variables $\mbox{\boldmath{$d$}}$. The presence or absence of the node $n_{j}$ in the data $d_{i}$ is indicated by the elements in eq.(\ref{BasketVector}).
\begin{eqnarray}
\mbox{\boldmath{$d$}}_{ij} = \left \{ \begin{array}{ll}
                    1 & \mbox{\ if $n_{j} \in d_{i}$} \\
                    0 & \mbox{\ otherwise}
                \end{array}
         \right . (0 \leq i < D, \ 0 \leq j < N).
\label{BasketVector}
\end{eqnarray}

Solving the node discovery problem means identifying all the surveillance logs where covert nodes would appear if they became overt. In other words, it means to identifying the logs for which $d_{i} \neq \delta_{i}$ holds because of the covert nodes belonging to $\mbox{\boldmath{$C$}}$.

\section{Solution}
\label{Solution}

\subsection{Heuristic method}
\label{Heuristic}

A heuristic method to solve the node discovery problem is studied in \cite{Mae09}. The method is reviewed briefly.

At first, every node which appears in the dataset $\{d_{i}\}$ is classified into one of the clusters $c_{l} \ (0 \leq l < C)$. The number of clusters is $C$, which depends on the prior knowledge. Mutually close nodes form a cluster. The measure of closeness between a pair of nodes is evaluated by the Jaccard's coefficient \cite{Lib04}. It is used widely in link discovery, web mining, or text processing. The Jaccard's coefficient between the nodes $n$ and $n'$ is defined by eq.(\ref{Jaccard}). The function $B(s)$ in eq.(\ref{Jaccard}) is a Boolean function which returns $1$ if the proposition $s$ is trueCor $0$ otherwise. The operators $\wedge$ and $\vee$ are logical AND and OR. 
\begin{eqnarray}
J(n,n') = \frac{\sum_{i=0}^{D-1} B( n \in d_{i} \wedge n' \in d_{i} ) }{ \sum_{i=0}^{D-1} B( n \in d_{i} \vee n' \in d_{i} ) }.
\label{Jaccard}
\end{eqnarray}

The k-medoids clustering algorithm \cite{Has01} is employed for classification of the nodes. It is an EM (expectation-maximization) algorithm similar to the k-means algorithm for numerical data. A medoid node locates most centrally within a cluster. It corresponds to the center of gravity in the k-means algorithm. The clusters and the modoid nodes are re-calculated iteratively until they converge into a stable structure. The k-medoids clustering algorithm may be substituted by other clustering algorithms such as hierarchical clustering or self-organizing mapping.

Then, suspiciousness of every surveillance log $d_{i}$ as a candidate where the covert nodes would appear is evaluated with a ranking function $s(d_{i})$. The ranking function returns higher value for a more suspicious log. The strength of the correlation between the log $d_{i}$ and the cluster $c_{l}$ is defined by $w(d_{i},c_{l})$ in eq.(\ref{Wdc}) as a preparation.
\begin{eqnarray}
w(d_{i},c_{l}) = \max_{n_{j} \in c_{l}} \frac{B(n_{j} \in d_{i})}{ \sum_{i=0}^{D-1} B(n_{j} \in d_{i}) }.
\label{Wdc}
\end{eqnarray}

The ranking function takes $w(d_{i},c_{l})$ as an input. Various forms of ranking functions can be constructed. For example, \cite{Mae09} studied a simple form in eq.(\ref{sd}) where the function $u(x)$ returns $1$ if the real variable $x$ is positive, or $0$ otherwise.
\begin{eqnarray}
s(d_{i}) &\propto& \sum_{l=0}^{C-1} u( w(d_{i},c_{l}) ) \nonumber \\
&=& \sum_{l=0}^{C-1} B(d_{i} \cap c_{l} \neq \phi).
\label{sd}
\end{eqnarray}

The $i$-th most suspicious log is given by $d_{\sigma(i)}$ where $\sigma(i)$ is calculated by eq.(\ref{Ranking2}). Suspiciousness $s(d_{\sigma(i)})$ is always larger than $s(d_{\sigma(i')})$ for any $i<i'$. 
\begin{eqnarray}
\sigma(i) = \arg \max_{m \neq \sigma(n) \ {{\rm for}} \ ^{\forall}n<i} s(d_{m}) \ (1 \leq i \leq D).
\label{Ranking2}
\end{eqnarray}

The computational burden of the method remains light as the number of nodes and surveillance logs increases. The method is expected to work generally for clustered networks but moderately even if the network topological and stochastic mechanism to generate the surveillance logs is not understood well. The method works without the knowledge about the hub-and-spoke model; the parametric form with $r_{jk}$ and $f_{j}$ in Section \ref{Problem}. The result, however, can not be very accurate because of the heuristic nature. A statistical inference method which requires heavy computational burden, but outputs more accurate results is presented next.

\subsection{Statistical inference method}
\label{Statistical}

The statistical inference method employs the maximal likelihood estimation to infer the topology of the network, and applies an anomaly detection technique to retrieve the suspicious surveillance logs which are not likely to realize without the covert nodes. The maximal likelihood estimation is a basic statistical method used for fitting a statistical model to data and for providing estimates for the model's parameters. The anomaly detection refers to detecting patterns in a given dataset that do not conform to an established normal behavior.

A single symbol $\mbox{\boldmath{$\theta$}}$ represent both of the parameters $r_{jk}$ and $f_{j}$ for the nodes in $\mbox{\boldmath{$O$}}$. $\mbox{\boldmath{$\theta$}}$ is the target variable, the value of which needs to be inferred from the surveillance log dataset. The logarithmic likelihood function \cite{Has01} is defined by eq.(\ref{Likelihood1}). The quantity $p(\{ d_{i} \}|\mbox{\boldmath{$\theta$}})$ denote the probability where the surveillance log dataset $\{ d_{i} \}$ realizes under a given \mbox{\boldmath{$\theta$}}.
\begin{eqnarray}
L(\mbox{\boldmath{$\theta$}}) = \log( p(\{ d_{i} \}|\mbox{\boldmath{$\theta$}})).
\label{Likelihood1}
\end{eqnarray}

The individual surveillance logs are assumed to be independent. eq.(\ref{Likelihood1}) becomes eq.(\ref{Likelihood2}).
\begin{eqnarray}
L(\mbox{\boldmath{$\theta$}}) &=& \log( \prod_{i=0}^{D-1} p(d_{i}|\mbox{\boldmath{$\theta$}})) \nonumber \\
&=& \sum_{i=0}^{D-1} \log( p(d_{i}|\mbox{\boldmath{$\theta$}})).
\label{Likelihood2}
\end{eqnarray}

The quantity $q_{i|jk}$ in eq.(\ref{fijk1}) is the probability where the presence or absence of the node $n_{k}$ as a responder to the stimulating node $n_{j}$ coincides with the surveillance log $d_{i}$. 
\begin{eqnarray}
q_{i|jk} = \left \{ \begin{array}{ll}
                    r_{jk} & \mbox{\ if $\mbox{\boldmath{$d$}}_{ik} = 1$ for given $i$ and $j$} \\
                    1-r_{jk} & \mbox{\ otherwise}
                \end{array}
         \right ..
\label{fijk1}
\end{eqnarray}

Eq.(\ref{fijk1}) is equivalent to eq.(\ref{fijk2}) since the value of $d_{ik}$ is either $0$ or $1$.
\begin{eqnarray}
q_{i|jk} = \mbox{\boldmath{$d$}}_{ik} r_{jk} + (1-\mbox{\boldmath{$d$}}_{ik})(1-r_{jk}).
\label{fijk2}
\end{eqnarray}

The probability $p(\{ d_{i} \}|\mbox{\boldmath{$\theta$}})$ in eq.(\ref{Likelihood2}) is expressed by eq.(\ref{Prob1}). \begin{eqnarray}
p(d_{i}|\mbox{\boldmath{$\theta$}}) = \sum_{j=0}^{N-1} \mbox{\boldmath{$d$}}_{ij} f_{j} \prod_{0 \leq k < N \ \wedge \ k \neq j} q_{i|jk}.
\label{Prob1}
\end{eqnarray}

The logarithmic likelihood function takes an explicit formula in eq.(\ref{Likelihood3}). The case $k=j$ in multiplication ($\prod_{k}$) is included since $d_{ik}^{2}=d_{ik}$ always holds.
\begin{eqnarray}
L(\mbox{\boldmath{$\theta$}}) = \sum_{i=0}^{D-1} \log ( \sum_{j=0}^{N-1} \mbox{\boldmath{$d$}}_{ij} f_{j} \prod_{k=0}^{N-1} \{ 1-\mbox{\boldmath{$d$}}_{ik} \nonumber \\
+(2\mbox{\boldmath{$d$}}_{ik}-1)r_{jk} \} ).
\label{Likelihood3}
\end{eqnarray}

The maximal likelihood estimator $\hat{\mbox{\boldmath{$\theta$}}}$ is obtained by solving eq.(\ref{Estimator}). It gives the values of the parameters $r_{jk}$ and $f_{j}$. A pair of nodes $n_{j}$ and $n_{k}$ for which $r_{jk} > 0$ possesses a link between them.
\begin{eqnarray}
\hat{\mbox{\boldmath{$\theta$}}} = \arg \max_{\mbox{\boldmath{$\theta$}}} L(\mbox{\boldmath{$\theta$}}).
\label{Estimator}
\end{eqnarray}

A simple incremental optimization technique; the hill climbing method (or the method of steepest descent) is employed to solve eq.(\ref{Estimator}). Non-deterministic methods such as simulated annealing \cite{Has01} can be employed to strengthen the search ability and to avoid sub-optimal solutions. These methods search more optimal parameter values around the present values and update them as in eq.(\ref{HillClimb}) until the values converge.
\begin{eqnarray}
\left \{ \begin{array}{l}
                    r_{jk} \rightarrow r_{jk} + \Delta r_{jk} \\
                    f_{j} \rightarrow f_{j} + \Delta f_{j}
         \end{array}
\right . (0 \leq j, k < N) .
\label{HillClimb}
\end{eqnarray}

The change in the logarithmic likelihood function can be calculated as a product of the derivatives (differential coefficients with regard to $r$ and $f$) and the amount of the updates in eq.(\ref{deltaL}). The update $\Delta r_{nm}$ and $\Delta f_{n}$ should be in the direction of the steepest ascent in the landscape of the logarithmic likelihood function. 
\begin{eqnarray}
\Delta L(\mbox{\boldmath{$\theta$}}) = \sum_{n,m=0}^{N-1} \frac{\partial L(\mbox{\boldmath{$\theta$}})}{\partial r_{nm}} \Delta r_{nm} + \sum_{n=0}^{N-1} \frac{\partial L(\mbox{\boldmath{$\theta$}})}{\partial f_{n}} \Delta f_{n}.
\label{deltaL}
\end{eqnarray}

The derivatives with regard to $r$ are given by eq.(\ref{Partiallikelihood1}).
\begin{eqnarray}
\frac{\partial L(\mbox{\boldmath{$\theta$}})}{\partial r_{nm}} &=& \sum_{i=0}^{D-1} [ f_{n} \mbox{\boldmath{$d$}}_{in} (2\mbox{\boldmath{$d$}}_{im}-1) \nonumber \\
&\times& \prod_{k \neq m} \{ 1-d_{ik} +(2\mbox{\boldmath{$d$}}_{ik}-1) r_{nk} \} \nonumber \\
&\div& \sum_{j=0}^{N-1} \mbox{\boldmath{$d$}}_{ij} f_{j} \prod_{k=0}^{N-1} \{ 1-\mbox{\boldmath{$d$}}_{ik}+(2\mbox{\boldmath{$d$}}_{ik}-1)r_{jk} \} ]. \nonumber \\
& &
\label{Partiallikelihood1}
\end{eqnarray}

The derivatives with regard to $f$ are given by eq.(\ref{Partiallikelihood2}).
\begin{eqnarray}
\frac{\partial L(\mbox{\boldmath{$\theta$}})}{\partial f_{n}} &=& \sum_{i=0}^{D-1} [ \mbox{\boldmath{$d$}}_{in} \prod_{k=0}^{N-1} \{ 1-\mbox{\boldmath{$d$}}_{ik}+(2\mbox{\boldmath{$d$}}_{ik}-1)r_{nk} \} \nonumber \\
&\div& \sum_{j=0}^{N-1} \mbox{\boldmath{$d$}}_{ij} f_{j} \prod_{k=0}^{N-1} \{ 1-\mbox{\boldmath{$d$}}_{ik}+(2\mbox{\boldmath{$d$}}_{ik}-1)r_{jk} \} ]. \nonumber \\
& &
\label{Partiallikelihood2}
\end{eqnarray}

The ranking function $s(d_{i})$ is the inverse of the probability at which $d_{i}$ realizes under the maximal likelihood estimator $\hat{\mbox{\boldmath{$\theta$}}}$. According to the anomaly detection technique, it gives a higher return value to the suspicious surveillance logs which are less likely to realize without the covert nodes. The ranking function is given by eq.(\ref{Ranking1}). 
\begin{eqnarray}
s(d_{i}) = \frac{1}{p(d_{i}|\hat{\mbox{\boldmath{$\theta$}}})}.
\label{Ranking1}
\end{eqnarray}

The $i$-th most suspicious log is given by $d_{\sigma(i)}$ by the same formula in eq.(\ref{Ranking2}).

\section{Test Dataset}
\label{Test}

\subsection{Network}
\label{Network}

Two classes of networks are employed to generate a test dataset for performance evaluation of the two methods. The first class is computationally synthesized networks. The second class is a real clandestine organization. 

\begin{figure}
\begin{center}
\includegraphics[scale=0.27,angle=-90]{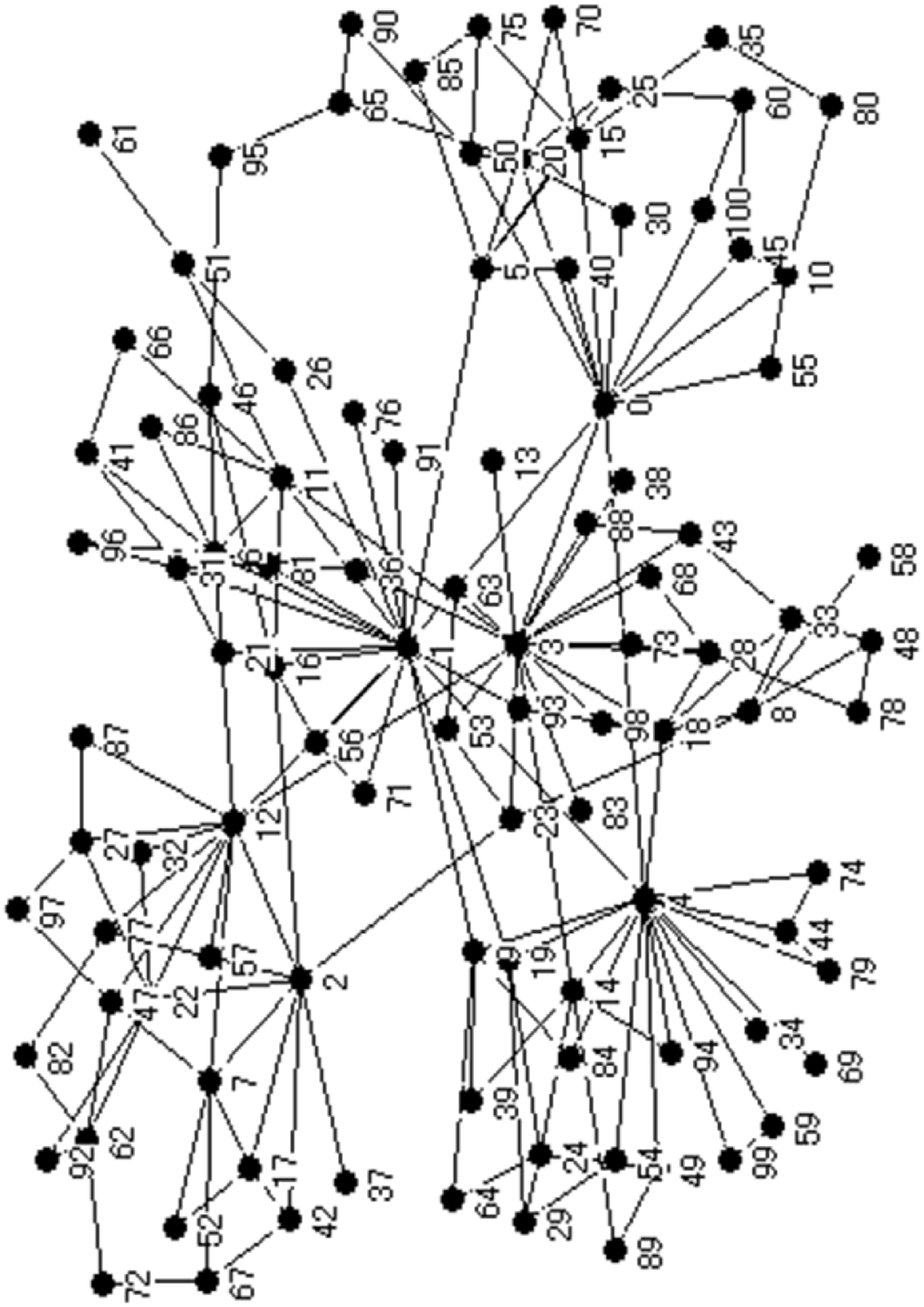}
\end{center}
\caption{Computationally synthesized network [A] which consists of $101$ nodes and $5$ clusters. Cluster contrast parameter is $\eta=50$. The network is relatively more clustered. The node $n_{{\rm 12}}$ is a typical hub node. The node $n_{{\rm 75}}$ is a typical peripheral node.}
\label{conn0902s}
%\end{figure}
%
%\begin{figure}
\begin{center}
\includegraphics[scale=0.27,angle=-90]{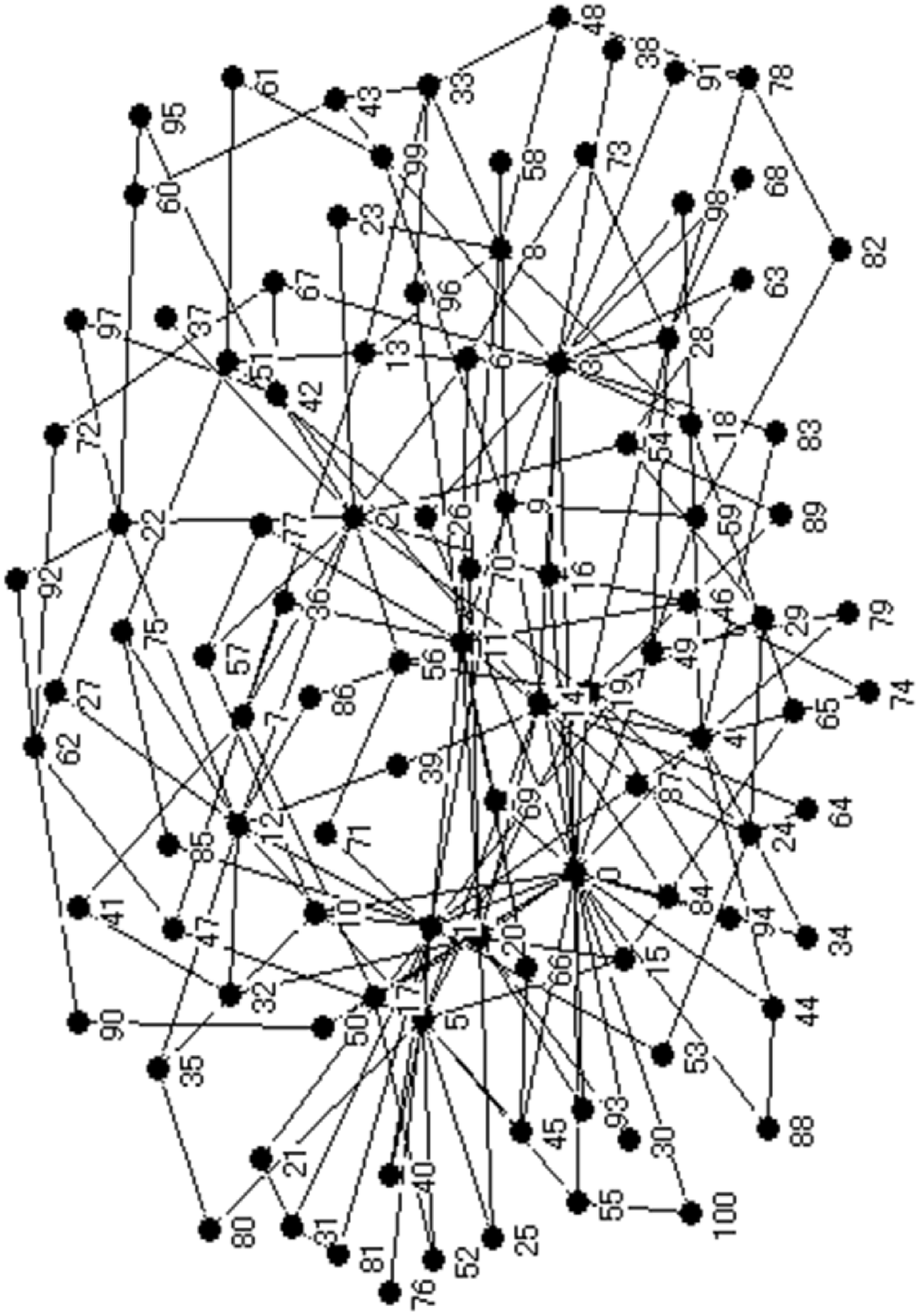}
\end{center}
\caption{
Computationally synthesized network [B] which consists of $101$ nodes and $5$ clusters. Cluster contrast parameter is $\eta=2.5$. The network is relatively less clustered. The node $n_{{\rm 12}}$ is a typical hub node. The node $n_{{\rm 48}}$ is a typical peripheral node.}
\label{conn0903s}
\end{figure}

The networks [A] in Figure \ref{conn0902s} and [B] in Figure \ref{conn0903s} are synthesized computationally. They are based on the Barab\'{a}si-Albert model \cite{Bar99} with a cluster structure. The Barab\'{a}si-Albert model grows with preferential attachment. The probability where a newly coming node $n_{k}$ connects a link to an existing node $n_{j}$ is proportional to the nodal degree of $n_{j}$ ($p(k \rightarrow j) \propto K(n_{j})$). The occurrence frequency of the nodal degree tends to be scale-free ($F(K) \propto K^{a}$). In the Barab\'{a}si-Albert model with a cluster structure, every node $n_{j}$ is assigned a pre-determined cluster attribute $c(n_{j})$ to which it belongs. The number of clusters is $C$. The probability $p(k \rightarrow j)$ is modified to eq.(\ref{preferential}). cluster contrast parameter $\eta$ is introduced. Links between the clusters appear less frequently as $\eta$ increases. The initial links between the clusters are connected at random before growth by preferential attachment starts.
\begin{eqnarray}
p(k \rightarrow j) \propto \left \{ \begin{array}{ll}
          \eta (C-1) K(n_{j}) & \mbox{if $c(n_{j}) = c(n_{k})$} \\
          K(n_{j}) & \mbox{otherwise}
        \end{array}
    \right .. \nonumber \\
\label{preferential}
\end{eqnarray}

Hub nodes are those which have a nodal degree larger than the average. The node $n_{{\rm 12}}$ in the network [A] in Figure \ref{conn0902s} is a typical hub node. Peripheral nodes are those which have a nodal degree smaller than the average. The node $n_{{\rm 75}}$ in the network [A] in Figure \ref{conn0902s} is a typical peripheral node.

\begin{figure}
\begin{center}
\includegraphics[scale=0.3,angle=-90]{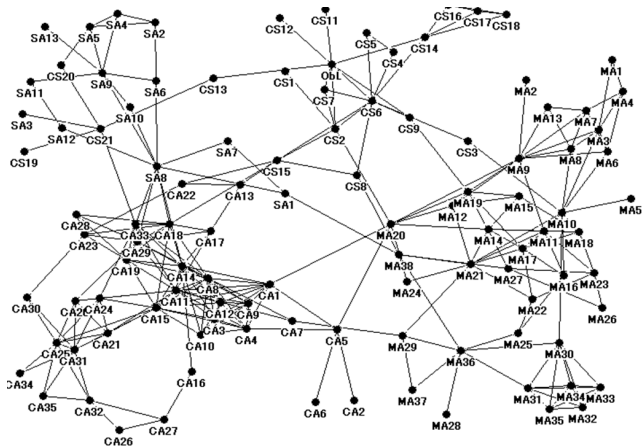}
\end{center}
\caption{Social network representing a global mujahedin (Jihad fighters) organization \cite{Sag04}, which consists of $107$ nodes and $4$ regional sub-networks. The sub-networks represent Central Staffs ($n_{{\rm CS}j}$) including the node $n_{{\rm ObL}}$, Core Arabs ($n_{{\rm CA}j}$), Maghreb Arabs ($n_{{\rm MA}j}$), and Southeast Asians ($n_{{\rm SA}j}$). The node $n_{{\rm ObL}}$ is Osama bin Laden who many believe is the founder of the organization.}
\label{conn0901s}
\end{figure}

The network in Figure \ref{conn0901s} represents a real clandestine organization. It is a global mujahedin organization which was analyzed in \cite{Sag04}. The mujahedin in the global Salafi jihad means Muslim fighters in Salafism (Sunni Islamic school of thought) who struggle to establish justice on earth. Note that jihad does not necessarily refer to military exertion. The organization consists of 107 persons and 4 regional sub-networks. The sub-networks represent Central Staffs ($n_{{\rm CS}j}$) including the node $n_{{\rm ObL}}$, Core Arabs ($n_{{\rm CA}j}$) from the Arabian Peninsula countries and Egypt, Maghreb Arabs ($n_{{\rm MA}j}$) from the North African countries, and Southeast Asians ($n_{{\rm SA}j}$). The network topology is not simply hierarchical. The 4 regional sub-networks are connected mutually in a complex manner.

The node representing Osama bin Laden; $n_{{\rm ObL}}$ is a hub ($K(n_{{\rm ObL}})=8$). He is believed to be the founder of the organization, and said to be the covert leader who provides operational commanders in regional sub-networks with financial support in many terrorism attacks including 9/11 in 2001. His whereabouts are not known despite many efforts in investigation and capture.

The topological characteristics of the above mentioned networks are summarized in Table \ref{TABLE1}. The global mujahedin organization has a relatively large Gini coefficient of the nodal degree; $G=0.35$ and a relatively large average clustering coefficient \cite{Wat98}; $\langle W(n_{j}) \rangle=0.54$. In economics, the Gini coefficient is a measure of inequality of income distribution or of wealth distribution. A larger Gini coefficient indicates lower equality. The values mean that the organization possesses hubs and a cluster structure. The values also indicate that the computationally synthesized network [A] is more clustered and close to the global mujahedin organization while the network [B] is less clustered.

\begin{table}
\caption{The number of nodes $M$, the number of clusters $C$, the average degree $\langle K(n_{j})\rangle$, the average clustering coefficient $\langle W(n_{j})\rangle$, and the Gini coefficient $G$ of the computationally synthesized networks (CSN) [A] and [B], and the global mujahedin organization (GMO).}
\begin{center}
\begin{tabular}{|c|cccccc|}
\hline
Model & $M$ & $C$ & $\eta$ & $\langle K \rangle$ & $\langle W \rangle$ & $G$ \\
\hline
CSN [A] & 101 & 5 & 50 & 3.6 & 0.42 & 0.36 \\
CSN [B] & 101 & 5 & 2.5 & 3.9 & 0.22 & 0.37 \\
GMO & 107 & - & - & 5.1 & 0.54 & 0.35 \\
\hline
\end{tabular}
\end{center}
\label{TABLE1}
\end{table}

\subsection{Test Dataset}
\label{Communication}

The test dataset $\{ d_{i} \}$ is generated from each network in \ref{Network} in the 2 steps below.

In the first step, the collaborative activity patterns $\{\delta_{i}\}$ are generated $D$ times according to the influence transmission under the true value of $\mbox{\boldmath{$\theta$}}$. A pattern includes both an initiator node $n_{j}$ and multiple responder nodes $n_{k}$. An example is $\delta_{{\rm ex}1}$ =$\{ n_{{\rm CS}1}$, $n_{{\rm CS}2}$, $n_{{\rm CS}6}$, $n_{{\rm CS}7}$, $n_{{\rm CS}9}$, $n_{{\rm ObL}}$, $n_{{\rm CS11}}$, $n_{{\rm CS12}}$, $n_{{\rm CS14}} \}$ for the global mujahedin organization in Figure \ref{conn0901s}.

In the second step, the surveillance log dataset $\{ d_{i} \}$ is generated by deleting the covert nodes belonging to $\mbox{\boldmath{$C$}}$ from the patterns $\{\delta_{i}\}$. The example $\delta_{{\rm ex}1}$ results in the surveillance log $d_{{\rm ex}1}$ = $\delta_{{\rm ex}1} \cap \overline{\mbox{\boldmath{$C$}}}$ = $\{ n_{{\rm CS}1}$, $n_{{\rm CS}2}$, $n_{{\rm CS}6}$, $n_{{\rm CS}7}$, $n_{{\rm CS}9}$, $n_{{\rm CS11}}$, $n_{{\rm CS12}}$, $n_{{\rm CS14}} \}$ if Osama bin Laden is a cover node; $\mbox{\boldmath{$C$}}$ = {$n_{{\rm ObL}}$}. The covert node in $\mbox{\boldmath{$C$}}$ may appear multiple times in the collaborative activity patterns $\{ \delta_{i}\}$. The number of the target logs to identify $D_{{\rm t}}$ is given by eq.(\ref{dt}).
\begin{eqnarray}
D_{{\rm t}} = \sum_{i=0}^{D-1} B(d_{i} \neq \delta_{i}).
\label{dt}
\end{eqnarray}

In the performance evaluation in Section \ref{Performance}, a few assumptions are made for simplicity. The probability $f_{j}$ does not depend on the nodes ($f_{j} = 1/M$). The value of the probability $r_{jk}$ is either 1 when a link is present between nodes, or 1 otherwise. It means that the number of the possible collaborative activity patterns is bounded. The influence transmission is symmetrically bi-directional; $r_{jk} = r_{kj}$.

\section{Performance}
\label{Performance}

\subsection{Performance measure}
\label{measure}

Three measures, precision, recall, and van Rijsbergen's F measure \cite{Kor97}, are used to evaluate the performance of the methods. They are commonly used in information retrieval such as search, document classification, and query classification. The precision $p$ is used as evaluation criteria, which is the fraction of the number of relevant data to the number of the all data retrieved by search. The recall $r$ is the fraction of the number of the data retrieved by search to the number of the all relevant data. The relevant data refers to the data where $d_{i} \neq \delta_{i}$. They are given by eq.(\ref{precision}) and eq.(\ref{recall}) They are functions of the number of the retrieved data $D_{{\rm r}}$. It can take the value from 1 to $D$. The data is retrieved in the order of $d_{\sigma (1)}$, $d_{\sigma (2)}$, to $d_{\sigma (D_{{\rm r}})}$. 
\begin{eqnarray}
p(D_{{\rm r}}) = \frac{\sum_{i=1}^{D_{{\rm r}}} B(d_{\sigma (i)} \neq \delta_{\sigma (i)})}{D_{{\rm r}}}.
\label{precision}
\end{eqnarray}
\begin{eqnarray}
r(D_{{\rm r}}) = \frac{\sum_{i=1}^{D_{{\rm r}}} B(d_{\sigma (i)} \neq \delta_{\sigma (i)})}{D_{{\rm t}}}.
\label{recall}
\end{eqnarray}

The F measure $F$ is the harmonic mean of the precision and recall. It is given by eq.(\ref{Fvalue}).
\begin{eqnarray}
F(D_{{\rm r}}) &=& \frac{1}{ \frac{1}{2} (\frac{1}{p(D_{{\rm r}})} + \frac{1}{r(D_{{\rm r}})}) } \nonumber \\
&=& \frac{2p(D_{{\rm r}})r(D_{{\rm r}})}{p(D_{{\rm r}})+r(D_{{\rm r}})}.
\label{Fvalue}
\end{eqnarray}

The precision, recall, and F measure range from 0 to 1. All the measures take larger values as the performance of retrieval becomes better.

\subsection{Comparison}
\label{Comparison}

The performance of the heuristic method and statistical inference method is compared with the test dataset generated from the computationally synthesized networks.

\begin{figure}
\begin{center}
\includegraphics[scale=0.25,angle=-90]{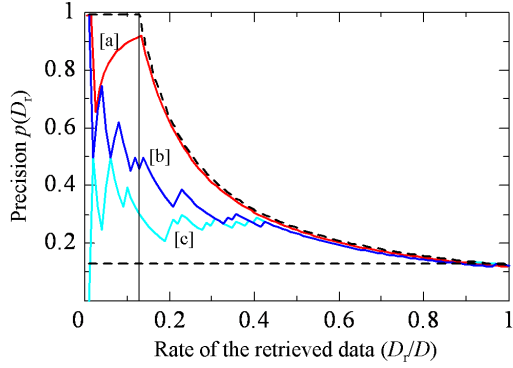}
\end{center}
\caption{Precision $p(D_{{\rm r}})$ as a function of the rate of the retrieved data to the whole data $D_{{\rm r}}/D$ in case the hub node $n_{12}$ in the computationally synthesized network [A] in Figure \ref{conn0902s} is the target covert node to discover. $\mbox{\boldmath{$C$}} = \{ n_{12}\}$. $|\mbox{\boldmath{$C$}}|=1$. $|\mbox{\boldmath{$O$}}|=100$. $D=100$. The three graphs are for [a] the statistical inference method, [b] the heuristic method ($C=5$), and [c] the heuristic method ($C=10$). The broken lines indicate the theoretical limit (the upper bound) and the random retrieval (the lower bound). The vertical solid line indicates the position where $D_{{\rm r}} = D_{{\rm t}}$.}
\label{conn0904s}
\end{figure}

\begin{figure}
\begin{center}
\includegraphics[scale=0.25,angle=-90]{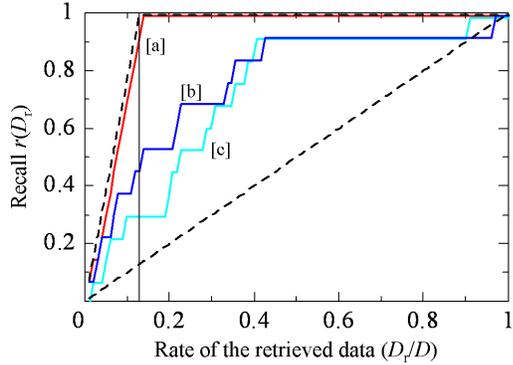}
\end{center}
\caption{Recall $r(D_{{\rm r}})$ as a function of the rate $D_{{\rm r}}/D$. The experimental conditions are the same as those for Figure \ref{conn0904s}.}
\label{conn0905s}
\end{figure}

\begin{figure}
\begin{center}
\includegraphics[scale=0.25,angle=-90]{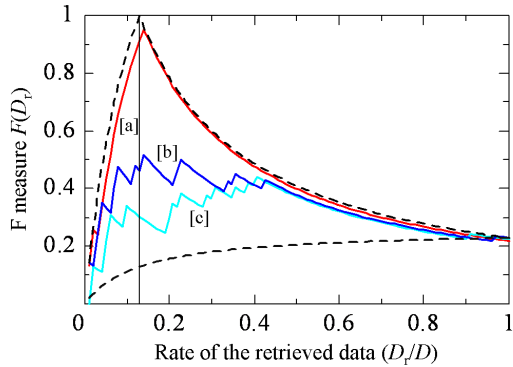}
\end{center}
\caption{F measure $F(D_{{\rm r}})$ as a function of the rate $D_{{\rm r}}/D$. The experimental conditions are the same as those for Figure \ref{conn0904s}.}
\label{conn0906s}
\end{figure}

Figure \ref{conn0904s} shows the precision $p(D_{{\rm r}})$ as a function of the rate of the retrieved data to the whole data $D_{{\rm r}}/D$ in case the hub node $n_{12}$ in the computationally synthesized network [A] in Figure \ref{conn0902s} is the target covert node to discover, $\mbox{\boldmath{$C$}} = \{ n_{12}\}$. The three graphs are for [a] the statistical inference method, [b] the heuristic method ($C=5$), and [c] the heuristic method ($C=10$). The number of the surveillance logs in a test dataset is $D=100$. The broken lines indicate the theoretical limit (the upper bound) and the random retrieval (the lower bound). The vertical solid line indicates the position where $D_{{\rm r}} = D_{{\rm t}}$. Figure \ref{conn0905s} shows the recall $r(D_{{\rm r}})$ as a function of the rate $D_{{\rm r}}/D$. Figure \ref{conn0906s} shows the F measure $F(D_{{\rm r}})$ as a function of the rate $D_{{\rm r}}/D$. The experimental conditions are the same as those for Figure \ref{conn0904s}. The performance of the heuristic method is moderately good if the number of clusters is known as prior knowledge. Otherwise, the performance would be worse. On the other hand, the statistical inference method surpasses the heuristic method and approaches to the theoretical limit. 

\begin{figure}
\begin{center}
\includegraphics[scale=0.25,angle=-90]{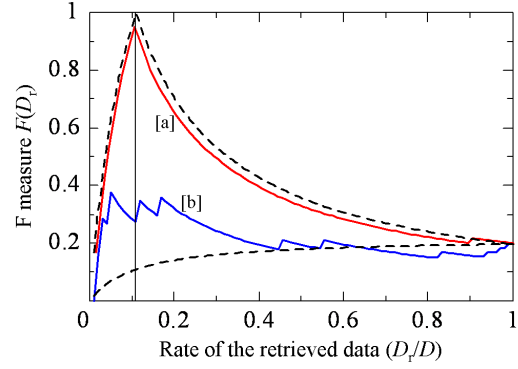}
\end{center}
\caption{F measure $F(D_{{\rm r}})$ as a function of the rate $D_{{\rm r}}/D$ in case the hub node $n_{12}$ in the computationally synthesized network [B] in Figure \ref{conn0903s} is the target covert node to discover. Two graphs are for [a] the statistical inference method, and [b] the heuristic method ($C=5$).}
\label{conn0907s}
\end{figure}

Figure \ref{conn0907s} shows the F measure $F(D_{{\rm r}})$ as a function of the rate $D_{{\rm r}}/D$ in case the hub node $n_{12}$ in the network [B] in Figure \ref{conn0903s} is the target covert node to discover. The two graphs are for [a] the statistical inference method and [b] the heuristic method ($C=5$). The performance of the statistical inference method is still good while that of the heuristic method becomes worse in a less clustered network.

\begin{figure}
\begin{center}
\includegraphics[scale=0.25,angle=-90]{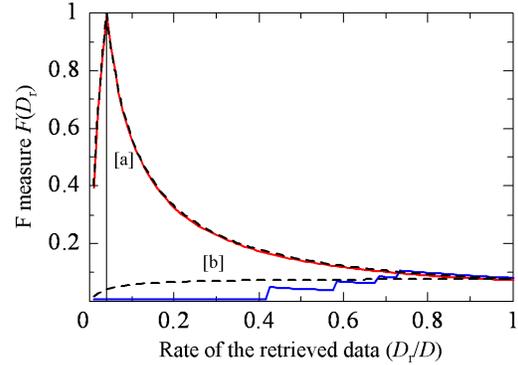}
\end{center}
\caption{F measure $F(D_{{\rm r}})$ as a function of the rate $D_{{\rm r}}/D$ in case the peripheral node $n_{75}$ in the computationally synthesized network [A] in Figure \ref{conn0902s} is the target covert node to discover. Two graphs are for [a] the statistical inference method, and [b] the heuristic method ($C=5$).}
\label{conn0908s}
\end{figure}

\begin{figure}
\begin{center}
\includegraphics[scale=0.25,angle=-90]{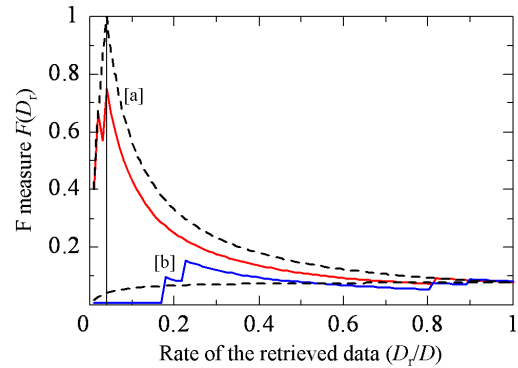}
\end{center}
\caption{F measure $F(D_{{\rm r}})$ as a function of the rate $D_{{\rm r}}/D$ when the peripheral node $n_{48}$ in the computationally synthesized network [B] in Figure \ref{conn0903s} is the target covert node to discover. Two graphs are for [a] the statistical inference method, and [b] the heuristic method ($C=5$).}
\label{conn0909s}
\end{figure}

Figure \ref{conn0908s} shows the F measure $F(D_{{\rm r}})$ as a function of the rate $D_{{\rm r}}/D$ in case the peripheral node $n_{75}$ in the network [A] in Figure \ref{conn0902s} is the target covert node to discover. Figure \ref{conn0909s} shows the F measure $F(D_{{\rm r}})$ as a function of the rate $D_{{\rm r}}/D$ when the peripheral node $n_{48}$ in the network [B] in Figure \ref{conn0903s} is the target covert node to discover. The statistical inference method works fine while the heuristic method fails.

\subsection{Application}
\label{Application}
I illustrate how the method aids the investigators in achieving the long-term target of the non-routine responses to the terrorism attacks. Let's assume that the investigators have surveillance logs of the members of the global mujahedin organization except Osama bin Laden by the time of the attack. Osama bin Laden does not appear in the logs. This is the assumption that the investigators neither know the presence of a wire-puller behind the attack nor recognize Osama bin Laden at the time of the attack.

\begin{figure}
\begin{center}
\includegraphics[scale=0.25,angle=-90]{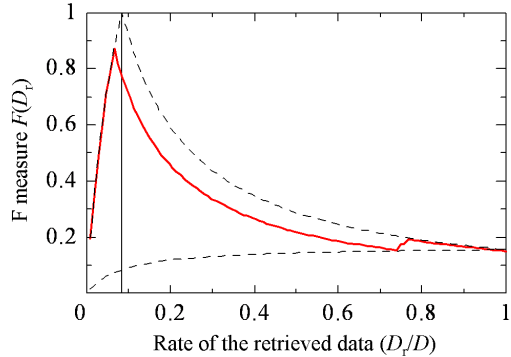}
\end{center}
\caption{F measure $F(D_{{\rm r}})$ as a function of the rate of the retrieved data to the whole data $D_{{\rm r}}/D$ when the statistical inference method is applied in case the node $n_{{\rm ObL}}$ in Figure \ref{conn0901s} is the target covert node to discover. $\mbox{\boldmath{$C$}}=\{n_{{\rm ObL}}\}$. $|\mbox{\boldmath{$C$}}|=1$. $|\mbox{\boldmath{$O$}}|=106$. The graph is for the statistical inference method. The broken lines indicate the theoretical limit and the random retrieval. The vertical solid line indicates the position where $D_{{\rm r}} = D_{{\rm t}}$.}
\label{conn0910s}
\end{figure}

The situation is simulated computationally like the problems addressed in \ref{Comparison}. In this case, the node $n_{{\rm ObL}}$ in Figure \ref{conn0901s} is the target covert node to discover, $\mbox{\boldmath{$C$}} = \{ n_{{\rm ObL}}\}$. Figure \ref{conn0910s} shows $F(D_{{\rm r}})$ as a function of the rate of the retrieved data to the whole data $D_{{\rm r}}/D$ when the statistical inference method is applied. The result is close to the theoretical limit. The most suspicious surveillance log $d_{\sigma(1)}$ includes all and only the neighbor nodes $n_{{\rm CS1}}$, $n_{{\rm CS2}}$, $n_{{\rm CS6}}$, $n_{{\rm CS7}}$, $n_{{\rm CS9}}$, $n_{{\rm CS11}}$, $n_{{\rm CS12}}$, and $n_{{\rm CS14}}$. This encourages the investigators to take an action to investigate an unknown wire-puller near these 8 neighbors; the most suspicious close associates. The investigators will decide to collect more detailed information on the suspicious neighbors. It may result in approaching to and finally capturing the covert wire-puller responsible for the attack.

The method, however, fails to identify two suspicious records $\delta_{{\rm fl}1}$=$\{ n_{{\rm ObL}}$, $n_{{\rm CS11}} \}$ and $\delta_{{\rm fl}2}$ = $\{ n_{{\rm ObL}}$, $n_{{\rm CS12}} \}$. These nodes have a small nodal degree; $K(n_{{\rm CS11}})=1$ and $K(n_{{\rm CS12}})=1$. This shows that the surveillance logs on the nodes having small nodal degree do not provide the investigators with much clues for the covert nodes.

\section{Conclusion}
\label{Conclusion}

In this paper, I define the node discovery problem for a social network and present methods to solve the problem. The statistical inference method employs the maximal likelihood estimation to infer the topology of the network, and applies an anomaly detection technique to retrieve the suspicious surveillance logs which are not likely to realize without the covert nodes. The precision, recall, and F measure characteristics are close to the theoretical limit for the discovery of the covert nodes in computationally synthesized networks and a real clandestine organization. In the investigation of a clandestine organization, the method aids the investigators in identifying the close associates and approaching to a covert leader or a critical conspirator.

The node discovery problem is encountered in many areas of business and social sciences. For example, in addition to the analysis of a clandestine organization, the method contributes to detecting an individual employee who transmits the influence to colleagues, but whose catalytic role is not recognized by company managers, may be critical in reorganizing a company structure. 

I plan to address two issues for the future works. The first issue is to extend the hub-and-spoke model for the influence transmission. The model represents the radial transmission from an initiating node toward multiple responder nodes. Other types of influence transmission are present in many real social networks. Examples are serial chain-shaped influence transmission model or tree-like influence transmission model. The second issue is to develop a method to solve the variants of the node discovery problem. Discovering fake nodes, or spoofing nodes are also interesting problems to uncover the malicious intentions of the organization. A fake node is the person who does not exist in the organization, but appears in the surveillance. A spoofing node is the person who belongs to an organization, but appears as a different node in the surveillance logs.

\onecolumn

\end{document}